\documentclass[runningheads]{llncs}
\usepackage[T1]{fontenc}
\usepackage{graphicx}
\usepackage{multirow}%
\usepackage{amsmath,amssymb,amsfonts}%

\usepackage{mathrsfs}%
\usepackage[title]{appendix}%
\usepackage{xcolor}%
\usepackage{textcomp}%
\usepackage{manyfoot}%
\usepackage{booktabs}%
\usepackage{algorithm}%
\usepackage{algorithmicx}%
\usepackage{algpseudocode}%
\usepackage{listings}%
\usepackage[font=small,labelfont=bf,tableposition=top]{caption}

\usepackage[linguistics]{forest}
\begin{document}

\title{EDC: Equation Discovery for Classification}
\author{Guus Toussaint\inst{1}\orcidID{0009-0008-7801-7212
} \and
Arno Knobbe\inst{1}\orcidID{0000-0002-0335-5099}}
\authorrunning{Toussaint, G. et al.}
\institute{
LIACS, Leiden University, Leiden, Netherlands\\
\email{\{g.toussaint,a.j.knobbe\}@liacs.leidenuniv.nl}}
\maketitle              % typeset the header of the contribution

\begin{abstract}
Equation Discovery techniques have shown considerable success in regression tasks, where they are used to discover concise and interpretable models (\textit{Symbolic Regression}). In this paper, we propose a new ED-based binary classification framework. Our proposed method EDC finds analytical functions of manageable size that specify the location and shape of the decision boundary. In extensive experiments on artificial and real-life data, we demonstrate how EDC is able to discover both the structure of the target equation as well as the value of its parameters, outperforming the current state-of-the-art ED-based classification methods in binary classification and achieving performance comparable to the state of the art in binary classification. We suggest a grammar of modest complexity that appears to work well on the tested datasets but argue that the exact grammar -- and thus the complexity of the models -- is configurable, and especially domain-specific expressions can be included in the pattern language, where that is required. The presented grammar consists of a series of summands (additive terms) that include linear, quadratic and exponential terms, as well as products of two features (producing hyperbolic curves ideal for capturing XOR-like dependencies). The experiments demonstrate that this grammar allows fairly flexible decision boundaries while not so rich to cause overfitting.
\end{abstract}



%%\pacs[MSC Classification]{35A01, 65L10, 65L12, 65L20, 65L70

\section{Introduction}

Equation discovery, the task of discovering analytical functions that model data, is well known for its use in regression settings, where it is then often referred to as Symbolic Regression \cite{billard2002symbolic,augusto2000symbolic}.
Many real-world applications require a model to be interpretable, particularly in scenarios where a decision could have substantial consequences.
Therefore, there is a movement in machine learning that tries to build models that can be easily interpreted and, where needed, corrected by the practitioners dealing with these models daily.
The framework of equation discovery fits this trend perfectly.
The result generated does not consist of a large, opaque model that calculates the value for the new data points but rather a simple equation that describes the nature of the data.
Furthermore, a domain expert can adjust the pattern language of equations considered by the algorithm, thus allowing the algorithm to adapt to different scenarios where, for example, a non-linear translation of the data makes sense.

As mentioned, the focus of equation discovery (ED) has largely been on regression problems.
In this work, we use the ED framework to tackle problems in the classification domain.
The central idea is to generate concise and readable equations that define the decision boundary between classes, which domain experts can then evaluate. 
We show that a classification-oriented ED algorithm may perhaps not beat the state-of-the-art of well-balanced algorithms such as Random Forests or Multi-Layer Perceptrons but be in the same ballpark in terms of classification performance and certainly produce more transparent models than, e.g. Random Forests would. 
Next to more interpretable models, we also show more expressive power where some standard mathematical functions may capture rather complex class distributions. 

When used in a regression setting, the equations considered are candidate functions to model the (numeric) target as a function of several input attributes. 
These equations typically involve arithmetic operators as well as a set of standard analytical functions ($\log$, $\exp$, $\sin$, ...). 
Such an approach could then be used to automate scientific discovery (such as Boyle's law, which relates the pressure of a gas in a vessel to its volume and temperature~\cite{physics}). 
However, the approach can also be used in more noisy statistical settings, where a regression model captures the influence of certain parameters on the target value. 
In broad strokes, ED has two separate challenges: first, finding the structure of the equation, and second, optimising the parameters of the candidate equation relative to the available data. 
Many standard optimisation algorithms are available for the latter, e.g., Gradient Descent \cite{GradientDescent}.
ED then primarily becomes a discrete optimisation problem defined by a search space of allowable equations and a strategy to traverse this search space, either exhaustively or heuristically. 
Like \cite{LjupcoTodorovski1997DeclarativeBI}, we define the pattern language, and hence the search space, by means of a configurable grammar that produces a language of equations, typically up to a certain complexity. 
We provide an example of a grammar that works quite well on datasets involved in our experiments, but we would like to stress that the exact grammar is a parameter of the proposed method. 
A knowledgeable domain expert may add certain functions to the grammar in order to enrich or reformulate the classification problem in terms of the functions provided.

We need a slightly different setting for ED-based classification since the target values are no longer numeric. 
Similar to Schwab et al.~\cite{symbolicregression1}, we model the decision boundary of the classifier in the following basic manner:
$T=true\ \textit{iff}\ f(x) >= 0$,
where $f(x)$ is the equation discovered by ED. 
In other words, the function $f$ defines a landscape in terms of the available attributes, and whenever the landscape is above 0, the classification is true. 
It is important to note that this paper focuses on \textit{binary classification}; extensions to multi-class classifications are possible but outside the scope of this paper.
For simple formulae, such as linear equations (hyperplanes), this setting is equivalent to existing methods such as logistic regression or SVMs with a linear kernel~\cite{LogisticRegression,flach}. 
However, ED comes into its own when richer expressions are involved.
% As a demonstration of the expressive power of only slightly more complex equations, consider the XOR task\footnote{Note that the original, stricter definition of an XOR function requires the outcome to be positive \emph{iff} exactly one of the signs is positive.} depicted in Figure \ref{fig:xor}. 
% Starting from the trivial function $c_0$ (which either predicts all cases as $true$ or $false$), within a single \emph{refinement}, we arrive at the following equation, which quite accurately models the data. 
% \[
%     c_0 + c_1 \cdot x_1x_2
% \]
% \noindent

% The equation has two parameters, $c_0$ and $c_1$, and a solution which results in an optimal separation of the classes is easily found with modern optimisers ($c_0 = 0$ and $c_1 = -7.81$), in this case, Gradient Decent.
% We require the found equations to be of limited complexity for several reasons, which include interpretability, computational explosion and risk of over-fitting.

% \begin{figure}[t]
%     \centering
%     \includegraphics[trim={0 0 0 0},clip,width=0.5\textwidth]{figures/xor_example.png}
%     \caption{Demonstration of a depth-1 equation that determines the depicted hyperbolic decision boundary. The underlying data is an artificial 2D XOR problem with cases being positive (black) \emph{iff} $x_0$ and $x_1$ have opposite signs.}
%     \label{fig:xor}
% \end{figure}

The contributions of our paper can be summarized as follows:
\begin{itemize}
    \item We propose a method called Equation Discovery for Classification (EDC), that is both interpretable and, with the correct building blocks for the domain at hand, is able to draw complex decision boundaries such that the performance is comparable to the current state-of-the-art in classification.
    \item With the aim of avoiding redundancy in the equation search space, we suggest a specifically-designed grammar that works in generic cases, and can be refined with application-specific constructs.
    \item We demonstrate that the algorithm is able to reconstruct a hard-coded decision boundary in artificial data and is able to model challenging XOR-like problems.
    \item We demonstrate the performance of our model on UCI datasets, where we compare it to state-of-the-art algorithms in terms of area under the curve (AUC), interpretability and model size.
\end{itemize}

Along with this paper, we publish the code\footnote{The code of the core algorithm can be found in the following repository: \url{https://github.com/GuusToussaint/EDC-core}.} and experimental setup\footnote{The code for reproducing the experiments can be found at the following link \url{https://github.com/GuusToussaint/EDC-experiments}.}\,\footnote{The datasets can be found at the following link: \url{https://anonymous.4open.science/r/EDC-datasets-B9F7/}.} so that the work presented here can be completely reproduced.

%%%%%%%%%%%%%%%%%%%%%%%%%%%%%%%%%%%%%%%%%%%%%%%%%%
\section{Related work}

The core idea of equation discovery, as it has been presented, is to find a function that fits the data.
To achieve this, a search strategy has to be defined; previous works broadly identify two types of search strategies: exhaustive and heuristic search.
Exhaustive search focuses on defining a manageable search space so that all possible equations within that search space can be evaluated.
One common approach to exhaustive search for equation discovery is to use a context-free grammar \cite{LjupcoTodorovski1997DeclarativeBI}. 
Implementing a heuristic search algorithm that traverses this search space more efficiently eliminates the need for restricting the initial search space.
Genetic programming has been successfully implemented as a heuristic approach in equation discovery \cite{augusto2000symbolic,haeri2017statistical,antonov2024functional}.
These come in two flavours.
Some try to optimise the structure and the parameters of the equation simultaneously, while others first try to find the structure, and optimise the parameters at a later stage. 

Regression is the dominant data mining task in equation discovery (in which case people speak of \emph{symbolic regression}). 
The aim here is to discover an analytical expression involving various arithmetic operations and closed-form functions (such as $log$, $sin$, etc.) that model a supervised dataset. The target expression typically exhibits a tree-shaped structure (which needs to be discovered) and involves one or more parameters (that need to be `fit' to the data). 
For a complete overview of the symbolic regression approach, we refer to the survey paper by Makke et al.~\cite{makke2023interpretablescientificdiscoverysymbolic}.
Beyond modelling explicit relationships, equation discovery has also been applied to uncovering latent monotonic features in time-series data~\cite{LMFD}.

% Start with an explanation of the inner workings of AMAXSC

In recent years, a handful of efforts have been made to apply the machinery of symbolic regression to the classification task, sometimes called Symbolic Classification (SC).
Schwab et al.~\cite{symbolicregression1} introduced the topic by describing a simple yet elegant alteration of the classic symbolic regression approach using genetic programming that can solve binary classification problems.
The algorithm, AMAXSC, produces a symbolic expression using Genetic Programming, which is then thresholded to produce a discrete decision boundary (much like in our setting).
While AMAXSC produces interpretable equations, the experiments show that the method struggles to discover concise and accurate models, often choosing to improve intermediate expressions by adding additional terms, rather than extensively optimizing the model's parameters.
This tendency can be mitigated by severely limiting the allowed operators, variables and functions to occur in the expressions.

To improve upon the initial design of AMAXSC and extend it to a multi-label classification setting, Ingalalli et al.~\cite{10.1007/978-3-662-44303-3_5} introduced the Multi-dimensional Multi-class Genetic Programming algorithm (M$_2$GP).
They argued that a single expression is not informative enough to solve multi-class classification problems.
Therefore, they use a multi-expression representation, where each sample is mapped to a custom number of dimensions $d$, where each dimension is a learned equation.
Thus, this approach can be seen as mapping any sample to a latent feature space using $d$ transformations.
The final classification is done by calculating the distance between the sample and the centroid of each class in the latent feature space.
The initial experiments showed promising results with performance comparable to state-of-the-art classification methods.
However, as a result of having an unspecified number of latent features, interpretability is compromised (which is understandable, given the challenges of multi-class classification).

The original method has since been improved by eliminating the need to manually select the number of dimensions $d$ and instead learning it as a parameter within the method.
M3GP introduced by Mu\~noz et al.~\cite{munoz2015m3gp} extends M$_2$GP by allowing the search process to progressively search for the optimal number of new dimensions that maximize the classification accuracy.
In the most recent development, La Cava et al.~\cite{LACAVA2019260} introduced M4GP, which improves upon M$_2$GP and M3GP by simplifying the program encoding, using advanced selection methods, and archiving solutions during the run.
In our experiments, we demonstrate that M4GP is not able to reach state-of-the-art performance on \emph{binary} classification in terms of model accuracy.

Korns et al.~\cite{Korns2018} have developed the Multilayer Disciminant Classification (MDC) algorithm, a computationally efficient alternative to the M$_2$GP.
However, the same performance and interpretability drawbacks remain.
Our approach aims to improve the interpretation and ease of use in binary classification by having only a single expression, while achieving competitive accuracy, comparable to the state-of-the-art in classification.

\section{Background}

Assume that we have a dataset $X$ consisting of numeric attributes.
A data point $x \in X$ consists of $d$ numeric features such that $x \in \mathcal{R}^d$.
When faced with a dataset that contains categorical features, one-hot encoding transforms them into a set of numeric 0,1-features.
As this may explode the number of features when dealing with high-cardinality categorical features, we take the pragmatic approach to group one-hot encoded features whose frequency of occurrence is smaller than or equal to $2\%$.

To evaluate the fitness of an equation, we use the Log Loss~\cite{LogLoss}, also known as the logistic loss or the cross-entropy loss.
% It is defined as the negative log-likelihood of a logistic model that returns probabilities over its inputs.
% The log loss function is defined as follows \cite{LogLoss}:
% \begin{equation}
%     L = - (Y log(\hat{Y} + (1 - Y)log(1-\hat{Y}))
%     \label{eq:logg_loss}
% \end{equation}
% \noindent
% Where $Y$ is the vector containing the labels, $y \in Y$ and $y \in {0, 1}$.
% Let $\hat{Y}$ be the vector of probabilities obtained from the model, representing the likelihood of each data point belonging to the true class.
Since the input for the Log loss function represents probabilities, only values between 0 and 1 are allowed.
Therefore, the output of the equation is passed through the logistic function before calculating the loss.

Once an equation is found, the results are ranked based on their probability scores.
This ranking is used to calculate the area under the receiver operating characteristic curve (AUC), which determines the final performance of the model. 
The equation does not assign direct labels to input data but rather a probability score \cite{flach}.
An appropriate threshold can be chosen based on the costs of mislabeling false cases as true and vice versa.
In this work, we will choose a threshold that maximizes the accuracy (misclassification costs are assumed to be equal).
This is done by selecting a threshold value that maximizes the true and false positives ratio.

%%%%%%%%%%%%%%%%%%%%%%%%%%%%%%%%%%%%%%
\section{Method}
\label{sec:method}

The proposed framework for equation discovery in a classification setting consists of two main components.
First, candidate equations are constructed during the \textit{search} step.
Second, the constants in these candidate equations are optimised during the \textit{optimisation} step.

\subsection{Search}
We will define the extent of the search space by means of a context-free grammar, further defined below.
However, it is safe to say that, in most cases, the defined search space will be too large to examine exhaustively. 
Even when the search depth is limited, with the growing number of attributes in the dataset, the number of possible equations will explode quickly. 
For this reason, we will adopt a heuristic search strategy based on \emph{beam search}, which provides an attractive balance between exploration and exploitation.

The notion of using a context-free grammar to define the search space of possible equations in equation discovery was first introduced by Todorovski et al.~\cite{LjupcoTodorovski1997DeclarativeBI}.
The overall search space is limited by declaring a grammar that describes all possible valid equations. 
However, the grammar can be defined so that all relevant equations remain in the constrained search space.
Ideally, a domain expert defines the grammar for each domain the algorithm applies to.
A domain expert can give insight into translation combinations of relevant features for the domain.

Our generic, application-independent grammar should include expressive and effective equations that are at the same time interpretable. It includes equations that consist of basic arithmetic operations, including addition, multiplication, and exponentiation. It is designed with an eye on limiting the level of redundancy in the search space such that two equations that are syntactically different but semantically equivalent are only considered once. As a simple example, $x + c\cdot y$ is equivalent to $x - c\cdot y$, since subtraction can be achieved by negating $c$, so the latter is not part of the grammar. Other opportunities for pruning the search space are explained below the definition of the grammar.

The grammar is defined as follows:
$G = (\mathcal{N}, \mathcal{T}, \mathcal{R}, \mathcal{S})$, which contains the following sets of symbols and operations:
\begin{itemize}
    \item $\mathcal{N}$ contains all non-terminal symbols.
    \item $\mathcal{T}$ contains all terminal symbols.
    \item $\mathcal{R}$ contains the rewrite rules in the form $A \xrightarrow{} \alpha$ where $A \in \mathcal{N}$ and $\alpha \in (\mathcal{N} \cup \mathcal{T})^*$.
    \item $\mathcal{S}$ contains the start symbols.
\end{itemize}

For our experiments, we have constructed the following grammar $G_s$:
\begin{itemize}
    \item $\mathcal{N}_s = \{V, B, X\}$
    \item $\mathcal{T}_s = \{x_1, x_2, c, +, \cdot, exp\}$ 
    \item $\mathcal{R}_s = \{ V \xrightarrow{} c\ |\ V + B $, \newline
    \hspace*{9mm} $B \xrightarrow{} c \cdot X \ |\ c \cdot \exp(c \cdot X) \ |\ c \cdot X \cdot X,$ \newline
    \hspace*{9mm} $X \xrightarrow{} x_1 \ |\ x_2$
    \}
    \item $\mathcal{S}_s = \{V$\}.
\end{itemize}
Note that this grammar only applies to datasets containing two features, i.e., $x_1$ and $x_2$; for our experiments, the non-terminal set $X$ is expanded to fit the number of features available in the dataset.
% Figure~\ref{fig:example_parse_trees} shows two examples of parse trees, which can be interpreted as equations generated with our proposed grammar.

When dealing with real-world data, the numerical inputs might have an arbitrary scale, which affects the range of the constants to optimize, something we would like to avoid.
For this reason, prior to actually starting the EDC algorithm, we normalise the data by means of linear scaling to the interval $[0,1]$.
% Effectively, each $x_i$ in our equations is replaced with the following scaled $x_i^*$:
% \[
%     x_i^* = \frac{x_i - min_{x_i}}{max_{x_i}-min_{x_i}}
% \]
% \noindent
% where $min_{x_1}$ and $max_{x_1}$ denote the smallest and largest value of feature $i$, respectively.
% When EDC terminates, the found optimal equation is then rewritten to account for this transformation, such that it can be applied to the original data.

Our choice of grammar is based on a number of design principles. 
First off, to keep the search space to a manageable size, we require our equation to be a set of summands: a sequence of building blocks that are connected through addition. 
As an additional benefit, this keeps derived equations similar to their ancestors: the derived equation just has a summand added and remains unchanged otherwise. 
Furthermore, we only use addition, not subtraction, between summands because each summand has a constant $c$ that the numeric optimiser can set to a negative number if necessary. 
As for the summands, we allow a linear term $x$, multiplication of two features $x\cdot y$ (note that this includes quadratic terms $x^2$), and exponentiation $\exp(x)$. 

For each such term, one or more constants $c_i$ are introduced in specific locations to further parameterize the equation.
For the $\exp$ function, we add a constant with multiplication `inside' since that cannot be rewritten into a simpler form.
Note that the addition of a constant inside $\exp$ is superfluous because that can be rewritten as follows: $ \exp(a+x) = \exp(a) \cdot \exp(x) = c\cdot \exp(x) $
% Finally, since $\exp$ is included, there is no need for $\log$.
% To understand this, note that the magnitude of the function doesn't matter for classification, just the sign.
% The $\exp$ function then introduces an exponential relation between one summand and the other, with the respective constants providing a weighting between the two.
% This comparison could have been achieved by instead taking the $\exp$ of the second, which demonstrates that the two functions play a comparable role, and $\log$ is thus redundant:
% \begin{align*}
%     & \log(x) < c \cdot y \Leftrightarrow \\
%     & \exp(\log(x)) < \exp(c \cdot y) \Leftrightarrow  \\
%     & x < \exp(c \cdot y) 
% \end{align*}

The grammar is not set in stone; thus, when deploying the proposed classifier in a domain with different characteristics, it is possible to design a different grammar with the required characteristics.
% \begin{figure}[t]
%    \centering
%    \begin{forest}
%  [V
%    [$c$ +]
%    [B +
%        [$c \cdot X^2$
%             [$x_1$]
%        ]
%    ]
%    [B
%        [$c \cdot X$
%             [$x_2$]
%        ]
%    ]
%  ]
% \end{forest}
% \hspace*{6mm}
% \begin{forest}
% [V
%    [$c$ +]
%    [B
%     [$c \cdot X \ \cdot $
%         [$x_1$]
%     ]
%     [$X$
%         [$x_2$]
%     ]
%    ]
%  ]
% \end{forest}
%    \caption{
%        Example of two parse trees derived from grammar $G_s$.
%        The produced strings, in our case equations, are $c + c \cdot x_1^2 + c \cdot x_2$ and $c + c \cdot x_1 \cdot x_2$.
%        }
%    \label{fig:example_parse_trees}
% \end{figure}

% Some general info about beam search and domains where it has been applied
While the equation language $\Lambda$, and thus the search space, is defined by the context-free grammar defined above, we use \emph{beam search} to heuristically traverse a relevant but limited portion of this search space. 
This search algorithm has successfully been applied in many mining settings, e.g., subgroup discovery~\cite{meeng2014rocsearch} or multi-label learning~\cite{kumar2013beam}.
% Figure~\ref{fig:beam_search_example} illustrates an example of beam search.
% \begin{figure}
%     \centering
%     \includegraphics[width=0.4\textwidth]{figures/beam_search_example.png}
%     \caption{Beam search for Equation Discovery. The figure shows four search levels, starting at the top with the root equation consisting of a single constant $c_0$. The refinement operator produces a collection of candidate equations indicated by the top triangle. The $w$ most promising equations of these (the beam) are then refined to the next level, and so on. The white triangles combined indicate the collection of candidates considered. The grey beam indicates the candidates selected for refinement. The total search space grows exponentially in $d$ and is much larger than the triangles indicated here. The red line shows the alternative best-first search, which may produce a less optimal result due to its more greedy nature.}
%     \label{fig:beam_search_example}
% \end{figure}
This approach builds equations within $\Lambda$ by iteratively adding summands to the equation, starting from a starting constant $c_0$. For example, the equation $c_0 + c_1\cdot x_1 + c_2\cdot \exp(c_3\cdot x_2)$ is found at search depth $d=2$ by refining the starting equation $c_0$ twice, as follows:
\begin{align*}
    & c_0 \\
    & c_0 + c_1\cdot x_1  \\
    & c_0 + c_1\cdot x_1 + c_2\cdot \exp(c_3\cdot x_2).
\end{align*}

\subsection{Optimisation}

Each equation encountered in the search space contains constants that need to be optimised for the target at hand. Our choice of optimiser is dependent on the structure of the equation.
If the equation is partially differentiable with respect to the constants $c_i$, we employ stochastic gradient descent (SGD) as our optimisation algorithm.
On the other hand, when the equation is not differentiable, which is the case for the exponentiation summand $\exp(c\cdot x)$, we employ a \emph{Hill Climber}-based approach.

The Hill Climber allocates a portion $f$ of its budget to sample random configurations. 
Subsequently, the top $k$ configurations are chosen as the starting points for the hill climber. 
The remaining budget of $n\cdot (1-f)$ evaluation is then evenly distributed among these top $k$ configurations, for $n\cdot (1-f)/2km$ evaluations each (where $m$ is the number of features).
Each evaluation involves taking a step of size $\alpha$ in both directions for each feature, i.e., moving up and down.
These potential steps are then evaluated, and the one that results in the most significant decrease in loss is selected as the next step.

\section{Experiments}
This section describes the experiments conducted to evaluate the EDC algorithm.
The experiments are divided into two parts.
% First, we identify the best optimiser for optimising the constants.
First, we evaluate the performance of the EDC algorithm under various conditions using a series of artificial datasets.
Second, the EDC algorithm is evaluated on a collection of UCI datasets and compared to other common classifiers.

\subsection{Artificial}

% Now that the best optimiser has been selected, and the hyperparameters have been set, we continue to test the actual EDC algorithm, beginning with more artificial datasets.
To evaluate the performance, we have developed four sets of artificial data, increasing in degree of difficulty.
Each set will evaluate the performance of the EDC algorithm under different conditions.
First, we evaluate the algorithm's performance when the target equation is within the search space, and no noise is introduced to the decision boundary.
Next, we introduce noise to the decision boundary to assess the resilience of our approach under more realistic conditions.
Then, we assess the performance of artificial datasets generated using a more complex grammar, where the decision boundary lies beyond the provided search space for the EDC algorithm.
Finally, artificial datasets are generated using random Gaussian clusters. This scenario tests whether adequate classifiers can be induced in the absence of an explicitly defined decision boundary.

%%%%%%%%%%%%%%%%%%%%%%%%%%%
\subsubsection{Within search space}
\label{sec:artificial_within_sp}

This scenario aims to evaluate the algorithm's performance in the most optimal, structured setting, i.e. the target decision boundary is within the search space, and no noise is added.
To achieve this, $100$ random equations are sampled from within the search space provided to the algorithm. 
These equations are then fitted with random constants to create an artificial dataset.
% Notice that this is similar to the approach taken in Section~\ref{sec:experiments_optimisers}; however, in this case, we don't provide the EDC with the structure of the target equation.

\subsubsection{Within search space with noise}
\label{sec:artificial_within_sp_with_noise}
In this scenario, equations are sampled from within the algorithm's search space, similar to the first scenario. 
However, noise is added to the output, making the decision boundary fuzzy.
We take the following approach to achieve this `fuzzy' decision boundary.
An equation is sampled from within the search space and fitted with random values for the constants.
Then, $2000$ data points are drawn uniformly from $[-10,10]\times[-10,10]$, and labelled true or false using the sampled equations as the decision boundary (see Figs \ref{fig:example_EDC_vs_ground_truth} and \ref{fig:example_EDC_vs_rich_ground_truth}).
After a label is assigned, the points are randomly perturbed using Gaussian noise according to distribution $\mathcal{N}(0,2^2)$ (a mean of $0$ and a standard deviation of $2$).
The AUC is calculated for both to compare the proposed decision boundary to the target.
Note that, due to the noise added to the data points after label assignment, the target decision boundary will not achieve an AUC of 1, which is expected.
% This experiment aims to provide insights into the resilience of the proposed algorithm. 
Since all real-world datasets are affected by noise, it is crucial to assess EDC's ability to identify a suitable decision boundary despite the presence of noise.

\subsubsection{Beyond search space with noise}
In this scenario, decision boundaries are sampled from a more complex grammar than the one available to EDC.
We have extended the original building blocks presented in section~\ref{sec:method} with the following:
\[
    c \cdot X^3,  \,\,\,\,   c \cdot X^4
\]
Only equations that contain at least one of these building blocks were selected during the random sampling.
Since the EDC algorithm is not provided with these extended building blocks, it must approximate the resulting decision boundary with the building blocks present in its search space.
As in the previous paragraph, noise is added to the data points to achieve a fuzzy decision boundary.
This experiment provides insights into EDC's performance when attempting to approximate equations outside its search space, which is important since, in real-world scenarios, some interactions present in the data might be missing as building blocks in the search space.

\begin{table}[t]
    \centering
    \begin{tabular}{r|cc}
        \toprule
         & \multicolumn{2}{c}{AUC} \\
        Dataset & EDC & Original DB \\
        \midrule
        Within search space             & 0.999 ($\pm0.00$) & 1.000 ($\pm0.00$)\\
        Within search space with noise  & 0.951 ($\pm0.03$) & 0.943 ($\pm0.03$)\\
        % Beyond search space&0.9987 ($\pm0.00$) & 1.0000 ($\pm0.00$)\\
        Beyond search space with noise  & 0.962 ($\pm0.03$) & 0.956 ($\pm0.03$)\\
        \bottomrule
    \end{tabular}
    \vspace{1mm}
    \caption{
        This table shows the results of the EDC algorithm on our sets of artificial datasets.
        The results are presented in terms of mean AUC with the standard deviation in parentheses.
    }
    \label{tab:artificial_results}
\end{table}

\subsubsection{Guassian clusters}
This last scenario creates $100$ datasets using a mixture of six randomly generated Gaussian clusters.
The labels are assigned by randomly assigning two Gaussian clusters to the \emph{true} class and the remaining four to the \emph{false} class.
No explicit target equation exists since the Gaussian intersection determines the dataset's decision boundary.
This experiment is designed to evaluate the performance of the EDC algorithm in a setting where the data generated does not come directly from an equation.
This setting is the closest to the real world of all the artificial experiments presented in this paper.
We compare the results on the $100$ generated datasets with the following algorithms.
First, we compare against the competing symbolic classification algorithms AMAXSC and M4GP.
Second, we compare against other interpretable classification algorithms, namely the decision tree (Tree) and the LDA algorithm.
Finally, we also compare against the state-of-the-art in terms of classification performance, namely the multi-layer perception (MLP), the random forest algorithm (RF), and the support vector machine (SVM) with a radial basis function kernel.
The results are presented in terms of the mean AUC with the standard deviation on the $100$ generated datasets.

\subsubsection{Results}

\begin{figure}[t]
\centering
\begin{minipage}{.43\textwidth}
    \centering
    \includegraphics[trim={0 0 0 0},clip,width=1\textwidth]{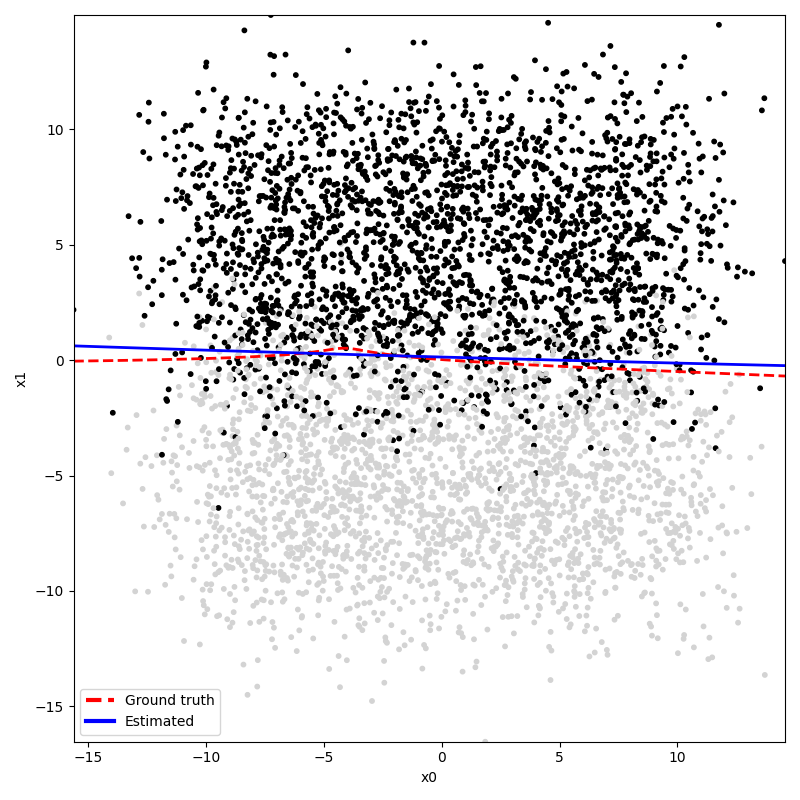}
    \caption{
    Target decision boundary (dashed red line), and decision boundary found by EDC (solid blue line).
    Noise is added to the data, and as a result, the target achieves a lower AUC ($0.952$) compared to the EDC algorithm ($0.978$).
    }
    \label{fig:example_EDC_vs_ground_truth}
\end{minipage}%
\hfill
\begin{minipage}{.43\textwidth}
    \centering
    \includegraphics[trim={0 0 0 0},clip,width=1\textwidth]{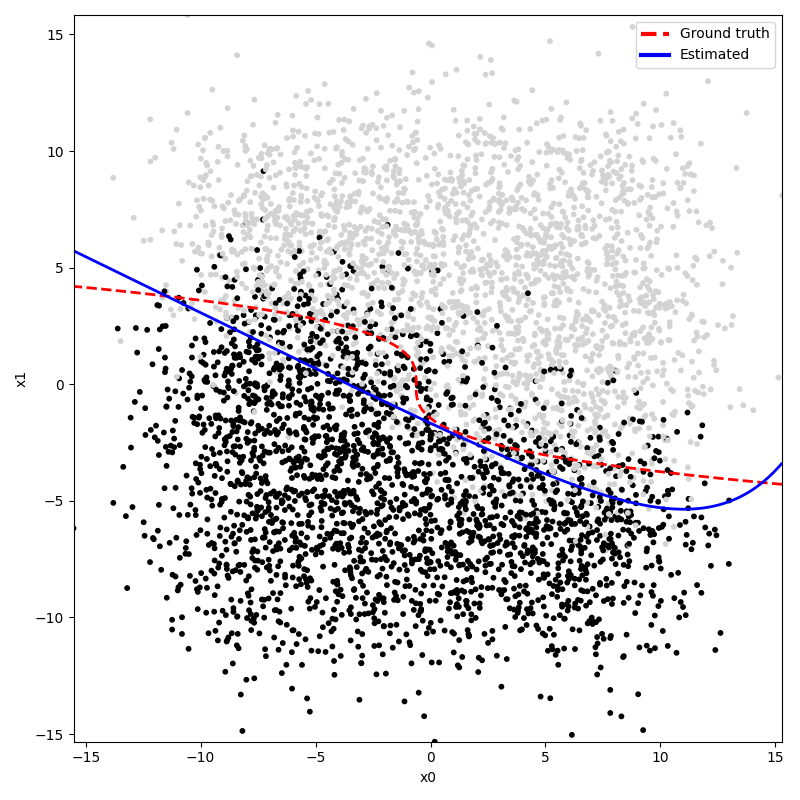}
    \caption{
    Target decision boundary (dashed red line), and decision boundary found by EDC (solid blue line).
    The target equation is sampled from a richer grammar than available to EDC.
    The target achieves a lower AUC ($0.959$) compared to EDC ($0.967$).
    }
    \label{fig:example_EDC_vs_rich_ground_truth}
\end{minipage}
\end{figure}

Table~\ref{tab:artificial_results} shows the results of our experiments on the artificial datasets.
We note that in the first setting, where the equation is from within the search space, and no noise is added to the decision boundary, the EDC algorithm is able to find approximate equations with an average AUC of $0.999$.
We can conclude that the EDC algorithm can identify the correct structure of the decision boundary. 
Any remaining error can likely be attributed to suboptimal optimisation of the constants.
However, we note that the obtained errors are very small.

When observing the results of a paired t-test for the second setting, we note a significant difference (in favour of EDC) between the original decision boundary (M = 0.9, SD = 0.03) and EDC (M = 1, SD = 0.03) AUC scores, with a $t(99) = 5.4$ and $p < .001$.
Similar behaviour is observed for the third setting, where the target (M = 1, SD = 0.03) and the EDC (M = 1, SD = 0.03) AUC scores also report a significant difference, $t(99) = 5.9$, $p < .001$, for the paired t-test.
Interestingly, when adding noise to the decision boundary, the EDC algorithm can find a new decision boundary that fits the data better than the original decision boundary.
We observe that for complex decision boundaries, the process of adding noise results in a new decision boundary being formed.
This is perhaps somewhat counter-intuitive since one would expect adding Gaussian random noise would not change the patterns hidden in the data.
Judging from the results, we note that the EDC algorithm can identify this \textit{new} decision boundary and subsequently achieves a marginal but significantly higher score than the original decision boundary in terms of AUC.
From the third setting, we observe similar behaviour, again showcasing that the EDC algorithm can identify a suitable decision boundary that significantly outperforms the original decision boundary in terms of AUC.
Figures~\ref{fig:example_EDC_vs_ground_truth} and \ref{fig:example_EDC_vs_rich_ground_truth} show examples of cases where the EDC algorithm outperforms the original decision boundary for the \textit{within search space with noise} and \textit{beyond search space with noise} datasets respectively.

\begin{figure}[t]
\centering
\begin{minipage}{.4\textwidth}
     \centering
    \includegraphics[width=1\textwidth]{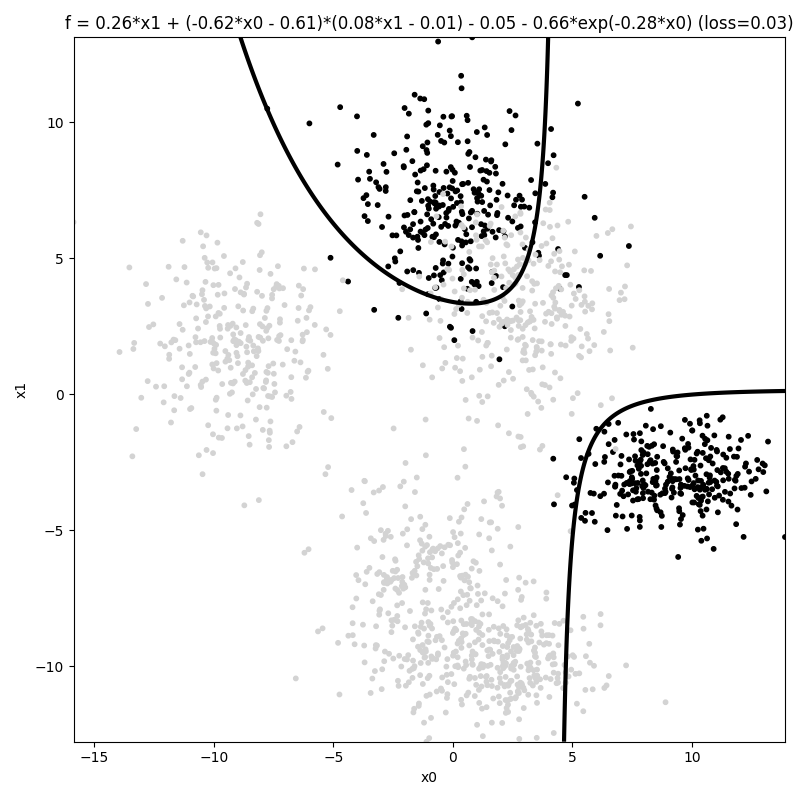}
    \caption{
        The proposed decision boundary for the Gaussian clusters artificial dataset. Note that the discovered equation indicated above the figure is produced after translating the equation back to the non-normalised space. This introduces two additional constants for each feature $x_i$.
    }
    \label{fig:DB_gaussian_clusters}
\end{minipage}%
\hfill
\begin{minipage}{.45\textwidth}
    \centering
    \small
    \begin{tabular}{r|c}
        \toprule
        Algorithm & AUC \\
        \midrule
        MLP &           0.972 ($\pm 0.034$)\\
        SVM &           0.970 ($\pm 0.038$)\\
        RF &            0.967 ($\pm 0.037$)\\
        \textbf{EDC} &  0.965 ($\pm 0.043$)\\
        Tree &          0.899 ($\pm 0.079$)\\
        M4GP &          0.894 ($\pm 0.082$)\\
        LDA &           0.811 ($\pm 0.139$)\\
        AMAXSC &        0.802 ($\pm 0.181$)\\
        \bottomrule
    \end{tabular}
    \vspace{2mm}
    \captionof{table}{
        This table shows the results of the artificial Gaussian clusters experiment.
        The results are presented in terms of mean AUC with the standard deviation in parentheses.
        The proposed EDC algorithm performs comparable to the state-of-the-art classification and outperforms all other explainable methods.
    }
    \label{tab:gaussian_comparison_results}
\end{minipage}
\end{figure}
 
Figure~\ref{fig:DB_gaussian_clusters} shows the proposed decision boundary for a single Gaussian clusters dataset.
We observe that the EDC algorithm is capable of defining a decision boundary that fits the data, even when the data-generating process is not guided by a \textit{target} decision boundary.
Table~\ref{tab:gaussian_comparison_results} shows the results of the comparison between the different classifiers.
The state-of-the-art classification algorithms perform the best, with mean AUC values around $0.97$.
Furthermore, we observe that the explainable methods represent the lower half of the ranking, with a notably high standard deviation across the $100$ generated datasets. This lower accuracy is likely caused by the limits on equation complexity introduced to promote explainability. In this respect, EDC appears to strike a good balance in the trade-off between accuracy and explainability. We note that EDC performs on par with the state-of-the-art and outperforms the existing explainable methods.

\subsection{UCI}

The final set of experiments in this paper involves the evaluation of the EDC algorithm on real-world datasets.
A list of UCI datasets that involve binary classification is selected.
% Table~\ref{tab:datasets} shows an overview of the statistics of the datasets selected in this work.
For each dataset, a 10-fold cross-validation approach is used, and the mean AUC scores (and standard deviations in parentheses) are reported.

% \begin{table}[t]
%     \centering
%     \begin{tabular}{r|cccc}
%         \toprule
%         Dataset     & Instances   & \#Numeric      & \#Nominal & \#Pre-proc. \\
%         \midrule
%         ADULT \cite{adult}       & 48 842	      & 6	        & 7	      & 49 \\
%         BANKNOTE \cite{misc_banknote_authentication_267}	& 1372 & 4	    & 0	      & 0 \\
%         BREAST \cite{breast_cancer_14}     & 286 & 0 & 10 & 39 \\
%         CREDIT \cite{credit_approval}	 & 690	  & 6           & 9	      & 35 \\
%         CYLINDER \cite{cylinder_bands_32}      & 378 & 20 & 19 & 77 \\
%         DIABETES	& 768	      & 8	        & 0	      & 0 \\
%         % HEPATITIS      & 83 & 6 & 13 & 26 \\
%         IONOSPHERE \cite{ionosphere}	& 351	      & 35	        & 0	      & 0 \\
%         OCCUPANCY \cite{misc_occupancy_detection__357} & 20 560 & 5 & 0 & 0 \\
%         SONAR \cite{sonar_uci}	    & 208	      & 60	        & 0	      & 0 \\
%         % WISCONSIN \cite{wisconsin_original}	& 699	      & 9	        & 0	      & 0 \\
%         \bottomrule
%     \end{tabular}
%     \vspace{1mm}
%     \caption{
%     Overview of UCI dataset details. Attribute counts exclude class attributes. All datasets concern binary classification. The \#Pre-proc. column indicates the number of attributes resulting from the one-hot encoding of nominal attributes and filtered on $p>0.02$.
%     }
%     \label{tab:datasets}
% \end{table}

For comparison to EDC, we have chosen the same set of algorithms as presented in Section~\ref{sec:artificial_within_sp}.

Table~\ref{tab:uci_results_normalized} shows the results of the different classifiers in terms of average AUC on the test set for 10-fold cross-validation.
When comparing the results of the EDC algorithm to the results of the other ED-based classifiers, we observe that for all datasets, the EDC algorithm achieves a higher AUC.
This shows that our approach outperforms the current state-of-the-art ED-based approaches on binary classification.
Furthermore, as noted in the introduction, the produced model, which is a single equation, is more interpretable than the results obtained from the M4GP algorithm.

To illustrate the interpretability of our approach, one of the equations from the ADULT dataset is shown here:
\begin{align*}
    0.75 - 1.27 \cdot \texttt{own-child} \cdot \texttt{education-num} + 3.37 \cdot \texttt{capitalgain} \\+ 8.01 \cdot \exp(8.18 \cdot \texttt{married-civ-spouse})
\end{align*}
Note that \texttt{own-child} and \texttt{married-civ-spouse} are essentially binary (from the one-hot encoding of nominal attributes). This makes the first summand a conditional negative term depending on \texttt{education-num} (a form of if-statement). Similarly, the last summand adds a constant term that depends on the binary value of \texttt{married-civ-spouse}: $8.01$ if F, and $8.01\cdot \exp(8.18) = 28,586$ if T.

When comparing the results to the other interpretable classification algorithms, we note that LDA achieves a higher score for $5$ of the $9$ datasets.
This indicates that these datasets contain a linear decision boundary.
We note that our EDC algorithm outperforms the simple decision tree algorithm in all datasets.
When comparing the results to the state-of-the-art classification methods, we observe that the random forest method outperforms our approach for $6$ of the $9$ datasets.
Furthermore, for the datasets IONOSPHERE and SONAR, the random forest, MLP, and SVM algorithms show substantially higher AUC scores compared to EDC.
This would indicate a relationship between the input features and class that is not present in our building blocks.

\begin{figure*}[t]
    \centering
    \includegraphics[trim={0 0 0 0},clip,width=1.0\textwidth]{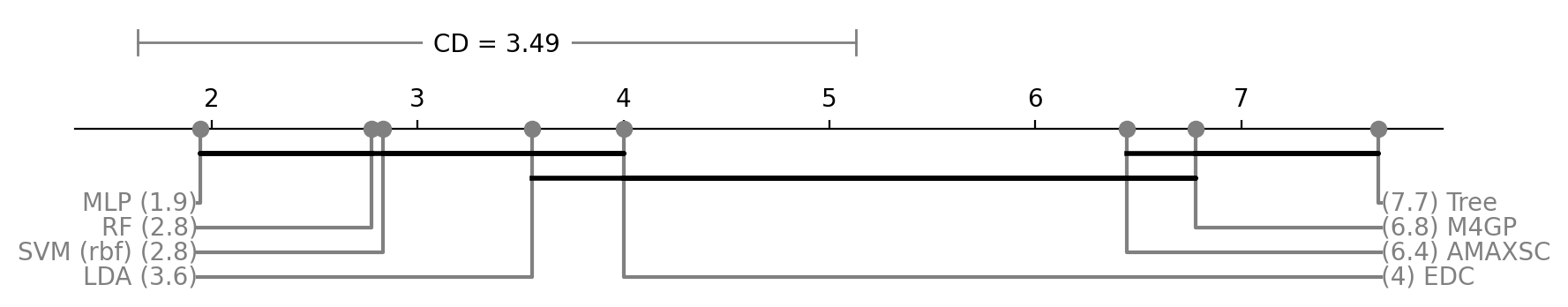}
    \caption{
        Critical distance plot of the ranks for the different classifiers for the UCI datasets.
        The top bar shows the critical distance ($CD$), which in our setup equals $3.49$. EDC outperforms AMAXSC, M4GP, and the decision tree, although not statistically significant. Similarly, MLP, RF, SVM, and LDA perform better but not statistically significantly.
    }
    \label{fig:critical_distance}
\end{figure*}

Figure~\ref{fig:critical_distance} shows a critical distance plot, as described by Demšar~\cite{JMLR:critical_difference}, of the ranks for the different classifiers evaluated in this work.
% The following equation obtains the value for the critical distance
% \[
% CD = q_\alpha \sqrt{\dfrac{k (k + 1)}{6N}}
% \]
% Where $k$ denotes the number of classifiers used in the comparison, which in our case is $8$, and $N$ denotes the number of datasets used in the comparison ($9$).
% The critical value $q_\alpha$ for $\alpha=0.05$ and $9$ classifiers equals $3.031$.
% Putting this together, we obtain the following $CD = 3.499$.
We identify three overlapping groups of classifiers: the first group consists of the MLP, RF, SVM, LDA and EDC algorithms.
The second group contains the LDA, EDC, AMAXSC, and M4GP algorithms.
The third and final group contains the AMAXSC, M4GP and decision tree algorithms.
This plot shows that the current state-of-the-art classifiers outperform the prior classification algorithms.
However, current state-of-the-art classification algorithms do not significantly outperform our proposed EDC algorithms. 

Finally, it is important to acknowledge the considerable run time involved with EDC, compared to the other algorithms.
The run time of the algorithm depends strongly on the number of examples in a dataset (in a linear fashion), as well as on the number of features.
% In particular, the building block that produces combinations of features ($c \cdot X\cdot X$) impacts the number of evaluated equations quadratically.
As an indication of the ballpark efficiency of EDC, the run time for the DIABETES dataset (a medium-size dataset) for the EDC algorithm is $\sim 167$ minutes.
% However, we note that while the time to obtain the model, i.e., the equation, takes longer, at inference time, only a simple equation needs to be evaluated to obtain the class of a new sample.

\begin{table}[t!]
    \centering
    \footnotesize
    \begin{tabular}{r|cccccccc}
        \toprule

        Dataset & \textbf{EDC} & AMAXSC & M4GP & LDA & Tree & MLP & RF & SVM\\
        \midrule
        
ADULT~\cite{adult} & 0.889 & 0.807 & 0.770 & 0.902 & 0.730 & 0.901 & 0.880 & 0.898\\
BANKNOTE~\cite{misc_banknote_authentication_267} & 1.000 & 0.982 & 0.999 & 1.000 & 0.979 & 1.000 & 1.000 & 1.000\\
BREAST~\cite{breast_cancer_14} & 0.670 & 0.617 & 0.614 & 0.636 & 0.590 & 0.701 & 0.683 & 0.709\\
CREDIT~\cite{credit_approval} & 0.918 & 0.896 & 0.869 & 0.924 & 0.812 & 0.910 & 0.935 & 0.920\\
CYLINDER~\cite{cylinder_bands_32} & 0.735 & 0.547 & 0.703 & 0.778 & 0.594 & 0.844 & 0.870 & 0.759\\
DIABETES & 0.830 & 0.799 & 0.724 & 0.829 & 0.673 & 0.843 & 0.826 & 0.836\\
IONOSPHERE~\cite{ionosphere} & 0.894 & 0.888 & 0.855 & 0.901 & 0.892 & 0.985 & 0.979 & 0.979\\
OCCUPANCY~\cite{misc_occupancy_detection__357} & 0.996 & 0.994 & 0.990 & 0.994 & 0.989 & 0.997 & 0.999 & 0.993\\
SONAR~\cite{sonar_uci} & 0.780 & 0.762 & 0.767 & 0.803 & 0.731 & 0.927 & 0.917 & 0.916\\
\midrule
Average Score & 0.857 & 0.810 & 0.810 & 0.863 & 0.777 & 0.901 & 0.899 & 0.890\\
Average Rank & 4.00 & 6.44 & 6.78 & 3.56 & 7.67 & 1.89 & 2.78 & 2.78\\

        \bottomrule
    \end{tabular}
    \vspace{1mm}
    \caption{
    Results of the classifiers on the UCI datasets.
    All scores are the mean Area Under the Receiver Operator Curve (AUC) across 10 folds.
    }
    \label{tab:uci_results_normalized}
\end{table}

% \begin{table*}[]
%     \centering
%     \begin{tabular}{c|cccc}
%     \toprule
%     Dataset & two step AUC & two step Hinge Loss & normal AUC & normal Hinge Loss \\
%     \midrule
%          ADULT &0.84 $\pm$ 0.01 &0.63 $\pm$ 0.01 &0.90 $\pm$ nan &0.87 $\pm$ nan  \\
% BANKNOTE &0.99 $\pm$ 0.01 &1.00 $\pm$ 0.00 &1.00 $\pm$ 0.00 &1.00 $\pm$ 0.00  \\
% BREAST &0.67 $\pm$ 0.07 &0.56 $\pm$ 0.08 &0.67 $\pm$ 0.11 &0.58 $\pm$ 0.10  \\
% CREDIT &0.92 $\pm$ 0.04 &0.89 $\pm$ 0.04 &0.92 $\pm$ 0.03 &0.90 $\pm$ 0.05  \\
% CYLINDER &0.67 $\pm$ 0.06 &0.59 $\pm$ 0.07 &0.73 $\pm$ 0.09 &0.69 $\pm$ 0.06  \\
% DIABETES &0.82 $\pm$ 0.04 &0.79 $\pm$ 0.05 &0.83 $\pm$ 0.04 &0.83 $\pm$ 0.05  \\
% HEPATITIS &0.65 $\pm$ 0.22 &0.77 $\pm$ 0.29 &0.68 $\pm$ 0.25 &0.89 $\pm$ 0.17  \\
% IONOSPHERE &0.86 $\pm$ 0.10 &0.88 $\pm$ 0.07 &0.89 $\pm$ 0.05 &0.92 $\pm$ 0.05  \\
% OCCUPANCY &1.00 $\pm$ 0.00 &0.99 $\pm$ 0.00 &1.00 $\pm$ 0.00 &0.99 $\pm$ 0.00  \\
% SONAR &0.73 $\pm$ 0.13 &0.72 $\pm$ 0.10 &0.78 $\pm$ 0.10 &0.78 $\pm$ 0.07  \\
% WISCONSIN &0.51 $\pm$ 0.09 &0.55 $\pm$ 0.11 &0.57 $\pm$ 0.11 &0.56 $\pm$ 0.13  \\
% \bottomrule
%     \end{tabular}
%     \caption{Caption}
%     \label{tab:my_label}
% \end{table*}

\section{Discussion \& Conclusion}
We have proposed a new classification algorithm based on equation discovery and thoroughly tested the method on artificial and real-life data (from the UCI Repository).
% Starting with the challenge of finding the optimal parameters for a given candidate equation and dataset, we compare different flavours of optimisation approaches, ranging from undirected random search to gradient-based approaches.
% It turns out that, within our suggested grammar of modest size, no optimisation approach is superior (nor inferior, for that matter), and a solution within $1\%$ of the optimum is always found.
% Apparently, even with up to 7 dimensions in the most complex equation, a thousand random function evaluations are sufficient to fit the equation.
% Still, it is expected that (due to the curse of dimensionality) with a deeper search and thus more complex equations involving more parameters, gradient descent methods will pull ahead from the others.

% Having established the adequate performance of the optimisation step, we moved on to test whether the heuristic search for equations in EDC was able to discover the correct equation and, thus, a successful decision boundary.

We evaluated whether the EDC algorithm was able to identify the correct equation on various artificial datasets, where the target equation is known; this was indeed the case.
In experiments where the target equation comes from the same grammar as EDC uses, AUC scores on datasets free of noise were in the order of $0.1\%$ from the perfect score.
Further experiments with artificial data that \emph{does} include noise showed an interesting phenomenon, where EDC statistically significantly outperformed the score of the target equation.
We deliberately avoid the term ‘ground truth’, since apparently, as soon as you introduce noise (data noise, not label noise), the implicit decision boundary embedded in the data may shift slightly from the original boundary. 
For linear decision boundaries, this is very limited and only due to the random sample.
However, in curved decision boundaries, it appears that the implicit boundary shifts towards the ‘enclosed’ region.
As a result, our algorithm was almost $1\%$ more accurate than the original equation.

Further experimentation with more challenging target equations outside EDC’s grammar (including cubic and quartic terms) demonstrated that EDC is able to approximate the intended decision boundary, again outperforming the original equation in the presence of noise. 
The results so far confirm that EDC’s beam search, which essentially ignores a large part of the discrete search space of equations, is able to focus on those candidate equations that are the most promising, at least on the artificial data. In fact, the current settings of beam width $w=10$ could perhaps be reduced in favour of shorter run times. 

It should be noted that the discovered equations, despite scoring well, may not necessarily take on the same structure as the original.
There apparently is some redundancy in our selected pattern language, such that different equations can still produce roughly the same decision boundary. 
Random effects in the sampling of the dataset or in the Gaussian noise allow some fuzziness in the boundary, which EDC sometimes exploits. 
For this reason, we have not included a grammatical validation between equations.

The final test concerned experiments on real-world data, for which typically no explicit decision boundary is known.
The results indicate that EDC is able to compete with state-of-the-art methods, such as MLPs and RFs, and outperforms all other ED-based classification methods.
Given EDC's demonstrated efficacy in a low-dimensional (artificial) dataset and its limited search depth, we hypothesise that EDC occasionally struggles when confronted with datasets that require a combination of many features, more than the maximum of six features our grammar imposes (e.g., $c_0 + c_1\cdot x_0x_1 + c_2\cdot x_2x_3 + c_3\cdot x_4x_5$). 
Still, the scores in Table \ref{tab:uci_results_normalized} demonstrate that this limitation does not frequently occur (most classification tasks apparently can be solved with relatively few features).

Finally, we must address the run time of the EDC algorithm.
As stated, this is significantly longer than the algorithms used in our comparison, which have run times of $< 1$ second.
The cause of this is the high number of function evaluations required in the search.
Especially the building block $x_i \cdot x_j$ introduces a quadratic number of building blocks.
This, combined with the feature expansion from one-hot encoding categorical features, results in long run times for the EDC algorithm.
% We should note that our selected hyper-parameters were not optimised for run time, so they are likely on the excessive side.
% For example, a more conservative beam width might lead to far fewer candidate equations being tested, possibly at a limited reduction in accuracy.
% Furthermore, we note that the number of optimisation steps, in our case $1000$, is excessive for some scenarios.
% A dynamic allocation of optimization steps could improve runtime without degrading the algorithm's overall performance.
% In future work, we will investigate this trade-off between run-time and the accuracy of EDC in more detail.
% Specifically, we expect to gain efficiency from reusing optimisation information from one equation to evaluating other similar equations.
%\vspace{-3mm}
\section*{Acknowledgements}
%\vspace{-3mm}
This work was financially supported by research project ``\emph{Ship system expanded energy storage devices lifetime via AI-empowered control (SEANERGETIC)}'' granted by Dutch research funding agency NWO.

\bibliographystyle{splncs04}
\bibliography{mybibliography}

\end{document}